\documentclass{article}

\usepackage{arxiv}

\usepackage[utf8]{inputenc} % allow utf-8 input
\usepackage[T1]{fontenc}    % use 8-bit T1 fonts
\usepackage{hyperref}       % hyperlinks
\usepackage{url}            % simple URL typesetting
\usepackage{booktabs}       % professional-quality tables
\usepackage{amsfonts}       % blackboard math symbols
\usepackage{nicefrac}       % compact symbols for 1/2, etc.
\usepackage{microtype}      % microtypography
\usepackage{graphicx}
\usepackage{doi}

\usepackage{subcaption}

\begin{document}

%%%%%%%%%%%%%%%%%%%%%%%%%%%%%%%%%%%%%%%%%%%%%%%%

% template ARXIV

\title{The Time Traveler's Guide to Semantic Web Research: Analyzing Fictitious Research Themes in the ESWC ``Next 20 Years'' Track} 

%\titlerunning{The Time Traveler's Guide to Semantic Web Research}

\author{\href{https://orcid.org/0000-0001-9962-7193}{\includegraphics[scale=0.06]{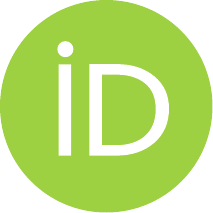}\hspace{1mm}Irene Celino}\\
Cefriel, viale Sarca 226, \\20126 Milano, Italy \\
\texttt{irene.celino@cefriel.com} 
\And 
\href{https://orcid.org/0000-0003-4386-819}{\includegraphics[scale=0.06]{orcid.pdf}\hspace{1mm}Heiko Paulheim}\\
University of Mannheim, B6 26, \\68159 Mannheim, Germany \\
\texttt{heiko.paulheim@uni-mannheim.de} 
}

% Uncomment to remove the date
\date{}

% Uncomment to override  the `A preprint' in the header
\renewcommand{\headeright}{}
\renewcommand{\undertitle}{}
\renewcommand{\shorttitle}{I. Celino and H. Paulheim -- The Time Traveler's Guide to Semantic Web Research}

%%%%%%%%%%%%%%%%%%%%%%%%%%%%%%%%%%%%%%%%%%%%%%%%

% template LNCS

%\title{The Time Traveler's Guide to Semantic Web Research: Analyzing Fictitious Research Themes in the ESWC ``Next 20 Years'' Track} 

%\titlerunning{The Time Traveler's Guide to Semantic Web Research}

%\author{Irene Celino\inst{1} \and Heiko Paulheim\inst{2}
%}

%\authorrunning{I. Celino and H. Paulheim}

%\institute{Cefriel, viale Sarca 226, 20126 Milano, Italy \\
%\email{irene.celino@cefriel.com} \\ \orcidID{0000-0001-9962-7193}
%\and University of Mannheim, B6 26, 68159 Mannheim, Germany \\
%\email{heiko.paulheim@uni-mannheim.de} \\ \orcidID{0000-0003-4386-819}
%}

%%%%%%%%%%%%%%%%%%%%%%%%%%%%%%%%%%%%%%%%%%%%%%%%%%%%%%%%%

\maketitle

\begin{abstract}
What will Semantic Web research focus on in 20 years from now? We asked this question to the community and collected their visions in the ``Next 20 years'' track of ESWC 2023. We challenged the participants to submit ``future'' research papers, as if they were submitting to the 2043 edition of the conference. The submissions -- entirely fictitious -- were expected to be full scientific papers, with research questions, state of the art references, experimental results and future work, with the goal to get an idea of the research agenda for the late 2040s and early 2050s. We received ten submissions, eight of which were accepted for presentation at the conference, that mixed serious ideas of potential future research themes and discussion topics with some fun and irony.

In this paper, we intend to provide a survey of those ``science fiction'' papers, considering the emerging research themes and topics, analysing the research methods applied by the authors in these very special submissions, and investigating also the most fictitious parts (e.g.,  neologisms, fabricated references). Our goal is twofold: on the one hand, we investigate what this special track tells us about the Semantic Web community and, on the other hand, we aim at getting some insights on future research practices and directions.
 

\end{abstract}

%\section{Paper structure}

%\begin{itemize}
%    \item why of the call
%    \item brief recap of the papers
%    \item insights from the authors (process they followed, characteristics of their submission, reflections they had or they wanted to solicit, ...)
%    \item insights from the papers (if we manage to have any analysis of language, word frequency, trends/topics, invented words, etc -- use spell checker for finding unusual words, double check for actual misspellings; create word cloud; identify trends from references)
%    \item insights from the audience (discussion at the conf, results of the survey, etc.)
%    \item considerations/speculations about the future of semweb research and research methodologies
%\end{itemize}

\section{Introduction}
%why of the call (some stuff from the proceedings preface)

The original paper by Tim Berners-Lee envisioning the Semantic Web in 2001 \cite{timbl2001semweb} included a future scenario in which structured knowledge and web technologies provided effective solutions to everyday problems. More than 20 years later, a lot changed and evolved in the Semantic Web community and, even if that \emph{exact} scenario did not become true as it was originally conceived, that vision has been the basic inspiration for our entire research field and the starting point for all the achievements in the Semantic Web realm and in related areas as well.

In its 2023 edition, The Extended Semantic Web Conference (ESWC) celebrated its 20th anniversary and during the conference there was the chance to reflect on what happened during the previous 20 years with a dedicated panel.\footnote{Cf. \url{https://2023.eswc-conferences.org/panel/}} The ESWC 2023 general chair -- Catia Pesquita -- % or the OC? or "we"?
% I have tried to describe concisely
decided to provide room to discuss not only the past, but also the future of the research in our community. For this reason, the ``Next 20 years'' Special Track was introduced. When appointed as chairs for this track, we first thought about inviting vision papers, but then decided to follow a more experimental route inviting \emph{future research papers} instead.

However, 20 years is quite a long period of time and it may be hard to predict what the actual trends will be within such a temporal span. Asking for envisioning the change over a period as long as 20 years is quite challenging. In a tongue-in-cheek-statement in 1999, science fiction author Douglas Adams said~\cite{adams1999howto}:
\begin{quote}
\begin{enumerate}
    \item \textit{Anything that is in the world when you’re born is normal and ordinary and is just a natural part of the way the world works.}
    \item \textit{Anything that’s invented between when you’re fifteen and thirty-five is new and exciting and revolutionary and you can probably get a career in it.}
    \item \textit{Anything invented after you’re thirty-five is against the natural order of things.}
\end{enumerate}
\end{quote}
Essentially, all attendees of ESWC 2023 will be over thirty-five in 2043\footnote{To the best of the authors' knowledge, there were no attendees aged 15 or below at ESWC 2023.}, so this task boils down to predict very radical new inventions and changes.
This is the reason why, in preparing the call for papers of the ``Next 20 years'' track, we decided to take a step forward into the future.

We were inspired by Design Fiction \cite{bleeker2022manual,bleeker2009designfiction}, a practice that mixes design, fiction, narratives, and speculations to create evocative ``artifacts'' that are aimed to represent the contexts and possible outcomes of change. As Bruce Sterling effectively explained \cite{sterling2013patently}, Design Fiction deliberately uses a set of prototypes (stories and artifacts that may support a story -- in our case scientific papers) in order to force an audience to ``suspend their disbelief'' about the future, thus being temporarily put in a different conceptual space (in our case, the future status of Semantic Web research).

Therefore, in the Call for Papers of the ``Next 20 years'' Special Track of ESWC, we invited the community to submit fictitious research papers, as if they were actually prepared for ESWC 2043. The papers were expected to look and feel like a real paper, including research questions, references, and state of the art (as of 2043), experimental results and possibly newer evaluation metrics, and of course future work (i.e., future future work), with the goal to get an idea of the research agenda for the late 2040s and early 2050s. We also encouraged the prospective authors to have their papers co-authored by an AI, thus imagining not only future research topics, but also future research practices.

To solicit an ``out of the box'' reflection, in the call for papers we also decided to illustrate the changes in the field over 20 years, by looking back at the past 20 years and at the key innovations brought forward by the Semantic Web research community. Some of those include:
\begin{itemize}
    \item OWL (2004) and SPARQL (2008) 
    \item FOAF (2004) and schema.org (2011)
    \item DBpedia (2007), Freebase (2007), Wikidata (2012)
    \item LOD Cloud (2007)
    \item R2RML (2012) and SHACL (2015)
    \item RESCAL (2011), TransE (2013), and RDF2vec (2016)
\end{itemize}
Considering how much we are used to these concepts, vocabularies, datasets, and techniques, some of them being referenced by hundreds and thousands of papers in the field, we asked the prospective authors to imagine a research landscape where upcoming inventions of similar impact have become standard textbook knowledge. 

\begin{figure}[htb]
     \centering
     \begin{subfigure}[b]{0.49\textwidth}
         \centering
         \includegraphics[width=\textwidth]{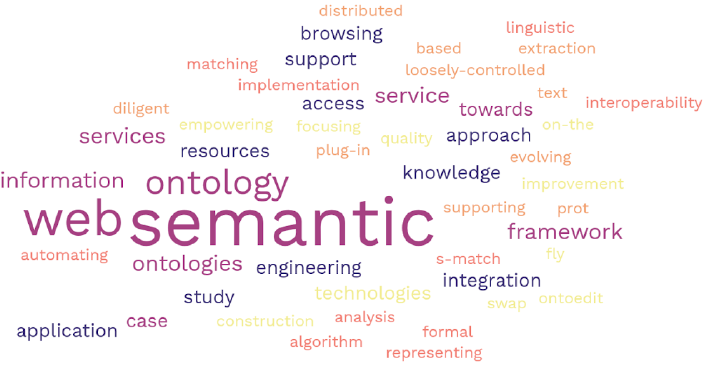}
     \end{subfigure}
     \hfill
     \begin{subfigure}[b]{0.49\textwidth}
         \centering
         \includegraphics[width=\textwidth]{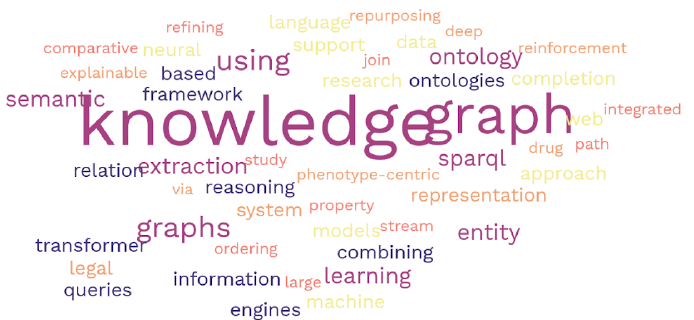}
     \end{subfigure}
    \caption{Word clouds compiled from accepted papers of the first edition of ESWC (left) and ESWC 2023 (right)}
    \label{fig:word_clouds}
\end{figure}

The changes over a couple of decades can also be visualized by having a look at the published research papers. For the field of Semantic Web research, Figure~\ref{fig:word_clouds} depicts word clouds compiled from the titles of accepted papers at the very first and the most recent edition of ESWC. 
We can easily spot that 20 years resulted in a shift of focus of the community: some words may have changed relevance (like ontologies, which are less represented), some others may have changed naming (less semantics and more knowledge), some expressions emerged (knowledge graphs is clearly a new popular buzzword), some others have disappeared  (like ``web'', maybe because of its ubiquity).

In this paper, we aim at summarising the results of this sort of ``social experiment'' that we conducted with the Semantic Web community, in an attempt to better understand the present and to draw some lines of the future to come. The remainder of the paper is structured as follows: in Section~\ref{sec:papers} we summarise the papers received and accepted to the track; in Section~\ref{sec:interviews} we provide some insights on how those papers were created, based on the interviews we conducted with the respective authors, while some analysis of the papers' content is offered in Section~\ref{sec:analysis}; the reaction and feedback of the ESWC 2023 audience to the papers presentation is illustrated in Section~\ref{sec:survey}, also with the results of the survey that we conducted; the challenges we encountered to publish this track's proceedings are explained in Section~\ref{sec:ceur}; finally, we present our final considerations in Section~\ref{sec:concl}.

%%%%%%%%%%%%%%%%%%%%%%%%%%%%%%%%%%%%%%%%%%%%%%%%%%%%%%%%%%%%%%%%%%%%%%%%%%%%%%%%%%%%%%%%%%%%
\section{The ``Future'' Papers}\label{sec:papers}
%brief recap of the papers (reference + a couple lines to describe their content)
The call for papers of this very special track\footnote{Cf. \url{https://2023.eswc-conferences.org/call-for-papers-the-next-20-years/}.} was published in the second half of February 2023, with the submission deadline extended until early April (less that 2 months overall). We also directly contacted some senior researchers of the community to solicit their visionary/futuristic submissions. In the end, we received 10 paper submissions and the two of us, as track chairs, peer reviewed them. After our discussion, we decided to accept 8 papers and to reject the other 2 papers, which were not in line with the call topic and spirit; among the accepted submission, we  distinguished 5 papers with a broader contribution for full presentation and 3 papers with a more limited scope for short presentation. All authors of accepted papers had thus the chance to present in a plenary session at the conference (cf. also Section~\ref{sec:survey}).

The accepted papers were quite varied in terms of proposition: there were both papers with a single author and with multiple authors, from a single institution (group-level contribution) or different organizations (cooperative contribution), from junior, senior or mixed groups of contributors, as summarised in Table~\ref{tab:summary}. We therefore think that the accepted papers are also representative of different possible research efforts, as in other more traditional conference tracks.
% We should soon have the proceedings ready and can link/reference them

\begin{table}[htb]
    \centering
    \begin{tabular}{l|c|c|c|c}
         & Authors & Seniority & Institution & Scope \\
         \hline 
        Anjomshoaa et al.~\cite{polleres2023wisdom} & group & mixed & single & broad \\
        Corcho et al.~\cite{corcho2023evoknomo} & group & mixed & single & broad \\
        d'Aquin~\cite{daquin2023omatch} & individual & senior & single & broad \\
        Wang~\cite{wang2023leveraging} & individual & junior & single & broad \\
        Motta et al.~\cite{motta2023robots} & group & mixed & cooperative & broad \\
        van Erp~\cite{vanerp2023pervasive} & individual & senior & single & specific \\
        Martorana et al.~\cite{martorana2023brainfreeze} & group & junior & single & specific \\
        Ilkou et al.~\cite{ilkou2023edumultikg} & group & junior & cooperative & specific \\
    \end{tabular}
    \caption{Summary of the eight accepted papers in terms of the involved authors, their seniority, their institutions and the paper scope.}
    \label{tab:summary}
\end{table}

The papers were also quite varied in terms of contents, even if all of them had some connection with relevant topics for the Semantic Web community: the fictitious journey towards a wisdom web with a universal language to ensure interoperability also at human level~\cite{polleres2023wisdom}, peer-to-peer knowledge sharing and its coherent evolution management in open data spaces opposed to large industry-closed efforts~\cite{corcho2023evoknomo}, agreement and matching between hyper-graph in a dystopian government-controlled ontology-mediated world~\cite{daquin2023omatch}, the evaluation of the contribution of ontologies to large language models that become ``standard''~\cite{wang2023leveraging}, the emergence of a new common sense and social norms in robot communities~\cite{motta2023robots}, different semantics emerging from brain waves measured through brain implants~\cite{martorana2023brainfreeze}, a very futuristic blending of the physical and the digital world with self-configuring materials and surfaces~\cite{vanerp2023pervasive}, and the successful application of knowledge graphs for user profiling in the education domain~\cite{ilkou2023edumultikg}.

While offering different views and inventing different possible worlds, those papers indeed address several broad topics, like the evolution of language (both for humans and machines), the dichotomy between centralized and decentralized data management, including the role of monopolist or quasi-monopolist organizations, the importance but also the vulnerability of personal data and knowledge, the opportunities of the convergence between symbolic and non-symbolic AI approaches, the relationship between humans and the rest of the world, with special emphasis on machines or AI-equipped objects. It is also worth noting that some papers clearly and explicitly provided specific reflections to their potential audience, either in the form of ``questions that people should have asked 20 years ago''~\cite{polleres2023wisdom} or ``check-list with the predictions'' to be evaluated in 2043~\cite{ilkou2023edumultikg}.

Finally, the papers are also quite diverse in terms of the degree of ``fictitiousness'': some authors really stretched their storytelling to include fantasy and unreal details, in order to push the boundaries of their design fiction exercise or simply to have fun, while others preferred to moderate their invented research to focus on real/realistic problems. It is worth noting that most papers included utopian, positive or neutral ``predictions'' about future events and scientific progress, in contrast to dystopian, negative or apocalyptic forecasts, which are also quite common in science fiction (see also Section~\ref{sec:survey}).

%%%%%%%%%%%%%%%%%%%%%%%%%%%%%%%%%%%%%%%%%%%%%%%%%%%%%%%%%%%%%%%%%%%%%%%%%%%%%%%%%%%%%%%%%%%%
\section{The Future Created by the Authors}\label{sec:interviews}
%insights from the authors (process they followed, characteristics of their submission, reflections they had or they wanted to solicit, ...)

In order to better understand how those papers were created, we conducted a series of interviews with the authors to collect some more information on the process they followed, the challenges they encountered and their overall experience. In this section, we provide an overview of the main qualitative aspects that emerged in response to our questions.

\subsection{Motivation to Participate}
We asked the authors why they decided to participate to this track. Most of them really liked the fun aspect of the call and they are also science fiction lovers, which made a good match. Many authors also underlined that they found the task challenging and were attracted by the novelty of the track: in this sense, they took the occasion to discuss at group level, brainstorming about current trends and challenges, collaboratively drawing interesting topics for their potential future agenda, to stretch the limits of usual discussions, to challenge themselves in a different intellectual exercise, to imagine their ideal future, to ask themselves serious questions about the future of the community, or simply just for fun.

\subsection{Process to Write the Paper}
We asked the authors what process they followed to write the paper and what similarities and differences they found with respect to their usual research methodology. It is worth noting that they provided different hints and reflections even when reporting similar practices, which is a sign that most of them did not consciously reflect on the method/process, which would therefore be interesting to further investigate and to compare both to traditional scientific paper writing and to science fiction production.

All authors reported to have followed a two-phase process, beginning with a brainstorming to find the focus/angle for their submission, followed by actually proceeding to the paper writing process. Those who organized group brainstorming were also interested to collect the potential ideas from a large number of potential contributors and to identify the most willing and suitable co-authors. Those who wrote single-authors papers motivated the choice mainly for time constraints, but in some cases also to be the unique creators of new worlds. In one single case, the authors, who are all PhD students, decided to ``go on their own'' without their respective supervisors also to challenge themselves to write a full paper without guidance.

The main difference that most authors reported with respect to their usual practices is that the brainstorming somehow continued also during the paper writing, in that some ideas were generated only during the manuscript creation, both because the process was not usual and because they found the need to fill some gaps in their storytelling only during composition. The result was that the final papers were quite different from the initial ideas (opposed to what happens when writing scientific papers, in which it is difficult that radical changes emerge during the writing phase). One author commented that they started ``really punky'', by inventing very futuristic and unrealistic scenarios, but closer to the submission deadline they decided to converge towards something more realistic. Some authors also used generative AI to support an iterative writing and refinement of the paper, either to entirely create the scenarios, contents and topics or to support the addition of specific parts or elements (see below).

The challenge to come up with entirely new or invented ideas was approached in different ways: some found it difficult, especially for the potential emergence of a large number of ideas that could have been difficult to reconcile; some others liked it exactly for the freedom and opportunity to go beyond the usual research conversations; some commented that the discussion within the group was more relaxed than conventional interactions between researchers. One author commented that inventing a future scenario is not that different from the usual creativity required in research to imagine the future, and that the discussion he had with his co-authors on this special occasion was very rewarding.

\subsection{Invented Details}
We also asked the authors to explain how they came up with a number of different details in their submission and if they employed generative AI.

All authors added to the papers their actual and current \emph{affiliation}; only a few of them additionally inserted some invented ones (e.g. a potential future NGO) and/or added AI co-authors and/or created an hypothetical co-author with a hypothetical future affiliation. Most of them said that they simply didn't think about changing their affiliation or that they would have found it weird to indicate a different institution, as the papers were going to be really published, even if they commented that it is not necessarily credible that they will still be working for the same institution 20 years from now. One author commented that he thought about changing his affiliation, but decided to leave his current contact information, to allow readers to contact him in case his submission would have generated further discussion or curiosity. Another author (a first year PhD student at a prestigious university) commented that leaving his current affiliation could have been also a wish for his own future career.

In some cases, the authors inserted \emph{invented words or names} (see also section~\ref{sec:analysis}): most authors simply created those by themselves, in some cases combining existing words in new expressions (e.g., \emph{wisdom web}, \emph{sensolens}, \emph{pervasive self-organization}, \emph{standard language models}); a few authors used ChatGPT to support the paper writing, mostly for editing, but in a few cases, this also supported the ``continuous brainstorming'', thus helping in coming up with new ideas, in creating neologisms (e.g. \emph{cognigram}) or in proposing the paper title. One author chose to make a strong use of Generative AI for the paper pictures (and actually she started her process from the images), also because she wanted to experiment with it, and only after that she started writing the paper, highlighting again the creativity-support role that this kind of technology can have on research paper production~\cite{dwivedi2023opinionchatgpt,oppenlaender2022creativity}. One paper also inserted a novelty in the paper title (as well in the titles of some of the citations, see below) by adding emoticons: the interviewed author motivated this as a reflection on how we will communicate research in the future. For another paper, the authors used ChatGPT to generate descriptions for scenarios they had sketched during brainstorming, with the idea to create a pipeline to generate/write multiple papers for different future scenarios; however, in the end they proceeded with a more traditional authoring process, because ChatGPT results were not truly original or coherent.

The papers that included an \emph{fictitious experimental evaluation} also provided figures and graphs to support their storytelling. The authors reported different perspectives on the task of inventing evaluation results: some found it easy (``we thought about what results we wanted to get, and then built the experiments and figures to support that''), some found it difficult (``at the beginning I thought it was an impossible task, but then one of my co-author did it and came up with very effective ideas, demonstrating that it was not that difficult after all''); some didn't like the experience (``I felt guilty to generate those numbers''), while some others enjoyed it and had fun (``I thought that it may be the only occasion in which I'm actually \emph{allowed} to invent perfect experimental results'').

One broad topic was the invention of \emph{citations of imaginary (future) related work}. Two completely different approaches were followed: manual crafting and automatic generation. The majority of authors decided to create the citations manually, to insert self-references and inventend future publications by their research groups or by other known researchers from the community (including in one case a paper by us, the track chairs) or to create papers referring to future events, in line with their storytelling. Some had fun in inventing authors (e.g. misspelled Chomsky) or in putting out-of-context names (e.g. cartoon characters, football players of the past); in one paper, a reference was inserted to satisfy the request by a colleague to have one paper ``published'' on a prestigious journal; in another case, the references were ordered from A to Z, with emoticons inserted in papers closer to 2043, to underline the possible change in scientific communication style. 
Some authors used generative AI to create fictional references, but then they reported that they had to fix them, because for example ``they were illogical with wrong years, so I had to change them manually to make the timeline consistent''. The challenge of straightening out a possible timeline of developments to put together a meaningful story of the future was a challenge reported by different authors.

Finally, we also observed different approaches for \emph{acknowledgements}: some authors mentioned current and existing institutions and funding, also because of their actual support to the paper writing; some other authors preferred not to insert any real acknowledgment, to avoid a potential disappointment of supporters not happy to be associated to a fake research work; in one case, the authors invented a future European-funded project on the topics covered by their paper; in another case, the author invented a fictional sponsor that supposedly provided the devices and supported the research, with a disclaimer saying that ``by contractual obligation, the author can only make positive statements'', with the intention to solicit the reflection on sponsored research, which is a topic highly discussed in domains like health, but mostly discarded in computer science.

\subsection{Plausibility of the Paper Content}
Finally we asked the authors to estimate what part of their papers could be considered plausible in terms of content and future predictions, or what part they would save if they were asked to write a ``serious'' vision paper on the future of the Semantic Web. We also asked about the kind of reflections they aimed to solicit in the audience. In this respect, the answers were really diverse, also reflecting the different characteristics of the papers, as summarised in Section~\ref{sec:papers}.

Some authors were mostly interested to discuss real topics and issues, affirming that more that 50\% of their papers could be considered realistic; for example, they focused on the shift from data spaces to knowledge spaces, or on the evolution of education, or on the need for new query languages, and they claim that those research trends are going to happen and it will be important to focus on them (even if, the actual realization may be different from their storytelling). Those authors believe that at least some of the vague ideas that they put on the papers are bound to happen in the future, and indeed they discovered that a few of the ideas came closer to reality after the paper submission, like the communication between LLM\footnote{Cf. \url{https://github.com/chatarena/chatarena}}.

Some other authors declared that they decided to give up on the feasibility/plausibility side at the beginning of their discussion, to feel free to focus on the fun side. In one case, the authors discovered after their submission that some idea they had included in the paper, thinking of it as crazy or unrealistic because related to interpretation of brain waves, was actually not that far from reality, as they found some similar concepts in papers~\cite{dado2023brain2gan} and news\footnote{Cf. \url{https://reut.rs/3sV09vH}.}. %\url{https://www.reuters.com/science/elon-musks-neuralink-gets-us-fda-approval-human-clinical-study-brain-implants-2023-05-25/}.}. 
In another case, the author affirmed: ``I think that my paper is 10\% plausible, but I cannot tell which part''.

Most authors declared their paper mostly or entirely non-plausible, but then -- being asked about ``what to save'' of their predictions -- all of them were able to identify an aspect, a detail or a topic that they find more realistic. It is the case for example of the risk of centralized knowledge controlled by companies, the evolution/standardization of language models, the stronger relationship between the physical and the digital worlds, the knowledge representation at individual level and the rise of data-intensive personal devices, the ``unintended consequences'' of evolutionary approaches, the issues of heterogeneity and the risk of under-representation of niche perspectives.

In most cases, as already mentioned in Section~\ref{sec:papers}, the papers touch upon clear and well-known topics of the Semantic Web community, because the authors asked themselves what our research field will actually become and what is needed to make it still relevant in 20 years. One author reported that he started by reflecting on what is actually going to change (like scalability of technology, data and devices) and then he built the paper, inventing a lot of fictional details around this idea. Another author said that they tried to exacerbate some realistic trends, also to counter-balance the current sensationalist debate on super-robots and AIs to become an existential threat to humanity. Other authors also explained that their intention was also to solicit a reflection on community building, stimulating a discussion on what the Semantic Web wants to become, especially to prevent the risk of being ``just a small AI community, which follows hypes and repeat old mistakes, instead of recognising and focusing on the central role of data and knowledge''.

One author reported that he had his paper reviewed by a friend who initially didn't understand the fictional aspect: he was excited by the ideas, tried to look up some of the mentioned research, was disappointed by failing to find them and complained about it; the author commented that this proved his goal of soliciting the reaction of ``I would like this research to actually happen''.

%%%%%%%%%%%%%%%%%%%%%%%%%%%%%%%%%%%%%%%%%%%%%%%%%%%%%%%%%%%%%%%%%%%%%%%%%%%%%%%%%%%%%%%%%%%%
\section{The Future Emerging from the Papers}\label{sec:analysis}
%insights from the papers (if we manage to have any analysis of language, word frequency, trends/topics, invented words, etc -- use spell checker for finding unusual words, double check for actual misspellings; create word cloud; identify trends from references)
Asides from asking the authors, we have also conducted a more systematic analysis of the papers. While a total number of eight papers does not allow for statistically significant solutions, we still were able to identify a few interesting findings.

\begin{figure}[ht]
   \centering
    \includegraphics[width=0.6\textwidth]{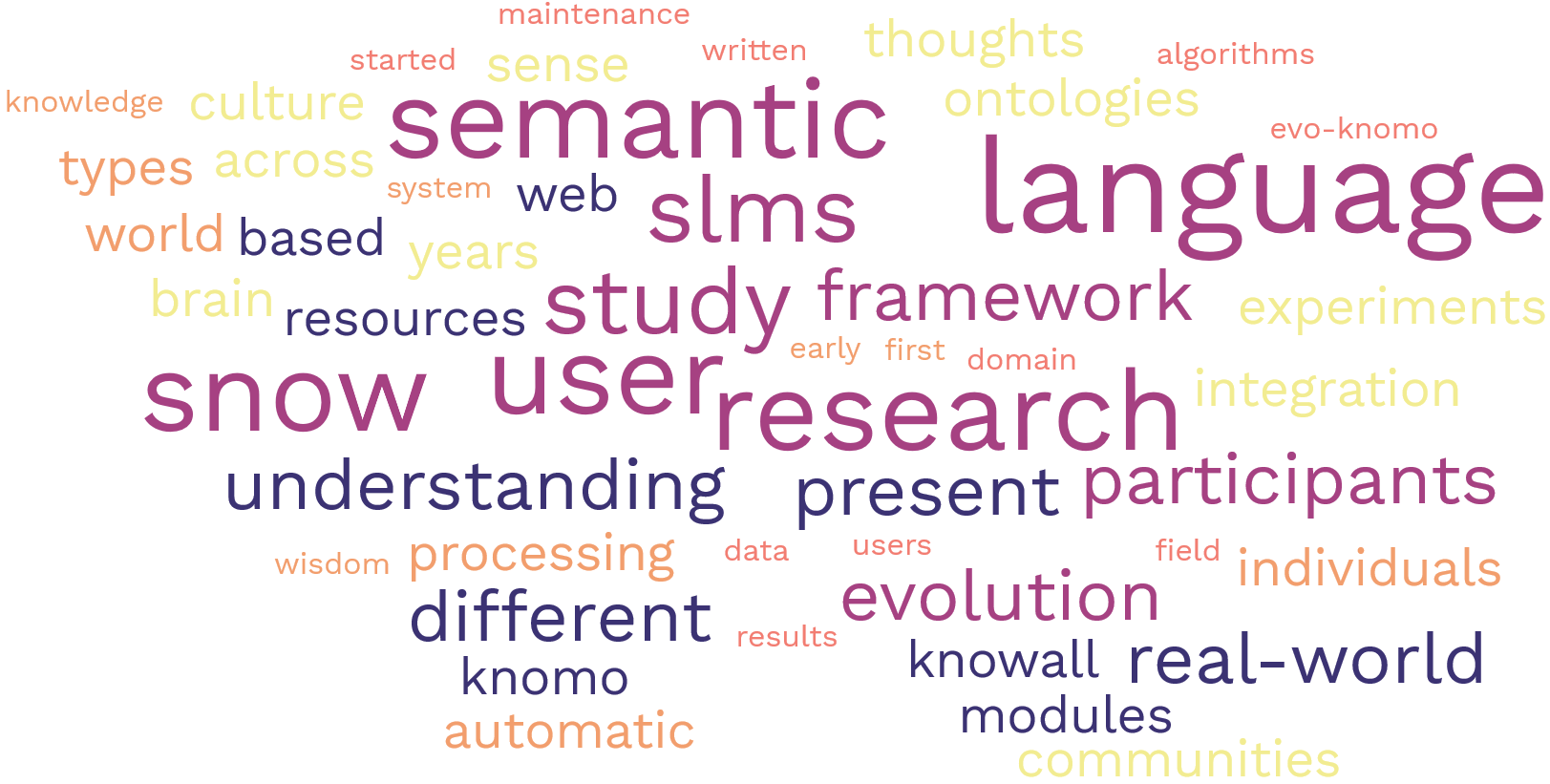}
    \caption{Word cloud from the abstracts}
    \label{fig:abstracts-word-cloud}
\end{figure}

Figure~\ref{fig:abstracts-word-cloud} shows a word cloud created from the papers' abstracts. We can see that \emph{semantics} and \emph{language} are commonly occurring words, the latter most often due to the use in \emph{language models}.\footnote{The strong appearance of the word ``snow'' is an artifact coming from only one paper mentioning this word rather often.}

To create a map of the research landscape created, we scanned the papers for abbreviations and constructed names\footnote{The papers were processed with regular expressions looking for words which contain at least two capitals (like \emph{SLM} or \emph{WisdomWeb}, or a capital not in the first position (like \emph{oMatch}).}. Those were double checked against existing terms to identify those describing future inventions. Table~\ref{tab:glossary} shows the glossary of future terms extracted using that method.

\begin{table}[ht]
    \footnotesize
    \centering
    \begin{tabular}{l|l|p{8cm}}
Name	&	Year	&	Description	\\
\hline
LGM~\cite{polleres2023wisdom}	&	2027	&	Large General Model, a structured knowledge equivalent to a large language model	\\
RXF~\cite{polleres2023wisdom}	&	2028	&	A knowledge exchange standard	\\
UniChip~\cite{corcho2023evoknomo}	&	2028	&	A small device for storing personal data	\\
CISQ~\cite{polleres2023wisdom}	&	2030s	&	An architecture for quantum computers, successor of NISQ	\\
QHedge~\cite{polleres2023wisdom}	&	2030s	&	Quantum hyper edge, a model for representing knowledge graphs on quantum computers	\\
SLM~\cite{wang2023leveraging}	&	2030s	&	Standard Language Model, a large language model enriched and regularized with ontologies	\\
WisdomWeb~\cite{polleres2023wisdom}	&	2030s	&	An infrastructure integrating data, LGMs, and personal AI assistants	\\
HAI~\cite{polleres2023wisdom}	&	2030s	&	Human-AI collaborative systems	\\
GoKnow~\cite{corcho2023evoknomo}	&	2030	&	A private knowledge management platform owned by Google	\\
HS~\cite{daquin2023omatch}	&	2032	&	Hierarchical Stance, a measure of agreement of a hyperedge and a graph	\\
Know4All~\cite{corcho2023evoknomo}	&	2033	&	A peer2peer knowledge management platform, open alternative to GoKnow	\\
ALD-UP~\cite{ilkou2023edumultikg}	&	2035	&	A user profiling algorithm in the educational domain	\\
Database2vec~\cite{wang2023leveraging}	&	2035	&	A vector space model for embedding databases	\\
UDF~\cite{polleres2023wisdom}	&	2037	&	Universal Description Framework, successor of RDF	\\
WikiMind~\cite{martorana2023brainfreeze}	&	2037	&	A Wikimedia projects for capturing thoughts	\\
iAGI~\cite{ilkou2023edumultikg}	&	2038	&	integrated Artificial General Intelligence	\\
MappingPedia2~\cite{corcho2023evoknomo}	&	2038	&	A system for sharing mappings on the KnoMo platform	\\
AdSkill~\cite{ilkou2023edumultikg}	&	2039	&	A user profiling algorithm in the educational domain	\\
EduKGs~\cite{ilkou2023edumultikg}	&	2039	&	Educational Knowledge Graphs	\\
OneWorld ontology~\cite{wang2023leveraging}	&	2039	&	An ontology of general world knowledge	\\
SmartClassroom~\cite{ilkou2023edumultikg}	&	2039	&	An IoT-based Learning Environment	\\
UL~\cite{polleres2023wisdom}	&	2039	&	Universal Language, a common language for AIs and Humans	\\
UTO~\cite{martorana2023brainfreeze}	&	2039	&	Universal Thought Ontology, an ontology of everything that can be thought	\\
XCon~\cite{motta2023robots}	&	2039	&	Obtaining explanations from an AI through conversation	\\
ConCensus~\cite{ilkou2023edumultikg}	&	2040	&	A consolidated census of the UN	\\
KnoMo~\cite{corcho2023evoknomo}	&	2040	&	Community-driven knowledge management platform 	\\
Q-Nets~\cite{wang2023leveraging}	&	2041	&	An AGI running on a quantum computer	\\
QGPT~\cite{motta2023robots}	&	2041	&	An Ultra Large Language Model	\\
ULLMs~\cite{motta2023robots}	&	2041	&	Ultra Large Language Models, successor of LLMs	\\
GST~\cite{motta2023robots}	&	2042	&	Generative Self-trained Transformer	\\
Prolearn-Ex~\cite{ilkou2023edumultikg}	&	2042	&	A user profiling algorithm in the educational domain	\\
BetaThink~\cite{martorana2023brainfreeze}	&	2043	&	A system for visualizing brain waves	\\
EvoKnoMo~\cite{corcho2023evoknomo}	&	2043	&	An approach for evolving KnoMo	\\
GPT-48, GPT-103~\cite{motta2023robots,vanerp2023pervasive}	&	2043	&	Sucessors of GPT-4	\\
Hyper-Reasoning Test~\cite{wang2023leveraging}	&	2043	&	A Benchmark for General Artificial Intelligence	\\
IAnepo~\cite{motta2023robots}	&	2043	&	A developer of language models	\\
MAWNED~\cite{ilkou2023edumultikg}	&	2043	&	Multimodal Adaptive Wisdom Network for Education	\\
oMatch~\cite{daquin2023omatch}	&	2043	&	A scalable system for sense integration	\\
SHReM~\cite{daquin2023omatch}	&	2043	&	Shared Hypergraph Relatedness Measure	\\
BarXiv~\cite{daquin2023omatch}	&	?	&	A blockchain-based version of arxiv	\\
ExplaPy~\cite{corcho2023evoknomo}	&	?	&	A Python library for explanations of AI systems	\\
HumOn~\cite{wang2023leveraging}	&	?	&	An ontology set for humanities	\\
Microsoft-NPIDIA~\cite{daquin2023omatch}	&	?	&	A technology vendor	\\
MindSynth~\cite{ilkou2023edumultikg}	&	?	&	A large pre-trained knowledge model	\\
ReinforceZ~\cite{motta2023robots}	&	?	&	A training algorithm for robots	\\
RoboGalleryOrg~\cite{vanerp2023pervasive}	&	?	&	An AI powered gallery organization system	\\
SciOn~\cite{wang2023leveraging}	&	?	&	An ontology set for science	\\
TrainE~\cite{motta2023robots}	&	?	&	A training algorithm for robots	\\
    \end{tabular}
    \caption{A glossary of future terms, extracted from the papers}
    \label{tab:glossary}
\end{table}

Common themes that can be observed from that gloassary are: standards and platforms for knowledge exchange (beyond RDF, also including integration across modalities such as knowledge graphs and LLMs), larger (and even ultra-large)  language models and those incorporating structured knowledge, quantum computers, and the extension of the graph model used in today's standards like RDF to hypergraphs. 

In order to analyze more deeply how the authors perceive and envision the current and upcoming trends in the research field, we also analyzed the references of the papers. In total, the eight papers contain 165 references, 28\% of which are from years up to 2023, the remaining 72\% from years 2024 onwards.

\begin{figure}[ht]
     \centering
    \includegraphics[width=0.6\textwidth]{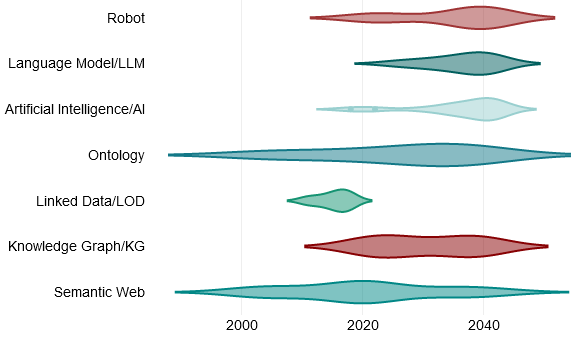}
    \caption{Key phrases from the references over time}
    \label{fig:reference-key-phrases-time}
\end{figure}

We extracted the key phrases and concepts from the references which (1) appear at least 5 times and (2) appear in the references of at least two submissions. Figure~\ref{fig:reference-key-phrases-time} shows an analysis of these keyphrases over time. They show a mix of concepts which are already there and will remain (ontologies, Semantic Web) and concepts which will gain importance (knowledge graphs, robots, large language models, artificial intelligence). Interestingly, none of the authors saw Linked (Open) Data play a crucial role in the next 20 years.

%new datasets

%new methods, metrics?

%%%%%%%%%%%%%%%%%%%%%%%%%%%%%%%%%%%%%%%%%%%%%%%%%%%%%%%%%%%%%%%%%%%%%%%%%%%%%%%%%%%%%%%%%%%%
\section{The Reception from the Conference Audience}\label{sec:survey}
%insights from the audience (discussion at the conf, results of the survey, etc.)

% some details on the presentations
On May 31st 2023, in a morning plenary session of ESWC 2023, the eight accepted papers were briefly presented by the respective authors. All presenters made an effort to make their storytelling entertaining, while making sure to convey the main ideas included in their paper. One of the authors decided also to explain the process he followed to write the paper, with an iterative process that involved ChatGPT to "fill some empty spaces" or for editing.

% some discussion of the Q&A afterwards
After the presentations, we had all speakers lined up in a panel setting to take questions from the audience. We did not have a concise plan on the Q\&A session and we did not give specific instructions on how and what to ask the authors. As a (partial) surprise to us, almost all questions were asked in the same spirit of the call, as if also the audience was in 2043, asking for clarifications or proposing alternative ideas. Some questions were made just for fun, but some others -- even if posed in an ironic way -- were aimed to tackle some specific relevant issue. Only one member of the community (Paul Groth) decided to ask a more ``serious'' question, inquiring about the motivations and goals they had in participating to the session, whether they were aimed to raise open issues and research directions or they just wanted to have fun, receiving quite different answers, as already explored in Section~\ref{sec:interviews}.

% results from the survey
At the end of the session, we asked the audience to compile a short survey to give us feedback about the track, the papers, the discussion and the future expectation. In total, 63 distinct users started the survey and 51 of them completed it (completion rate over 80\%). 

\begin{figure}[ht]
    \centering
    \includegraphics[width=\textwidth]{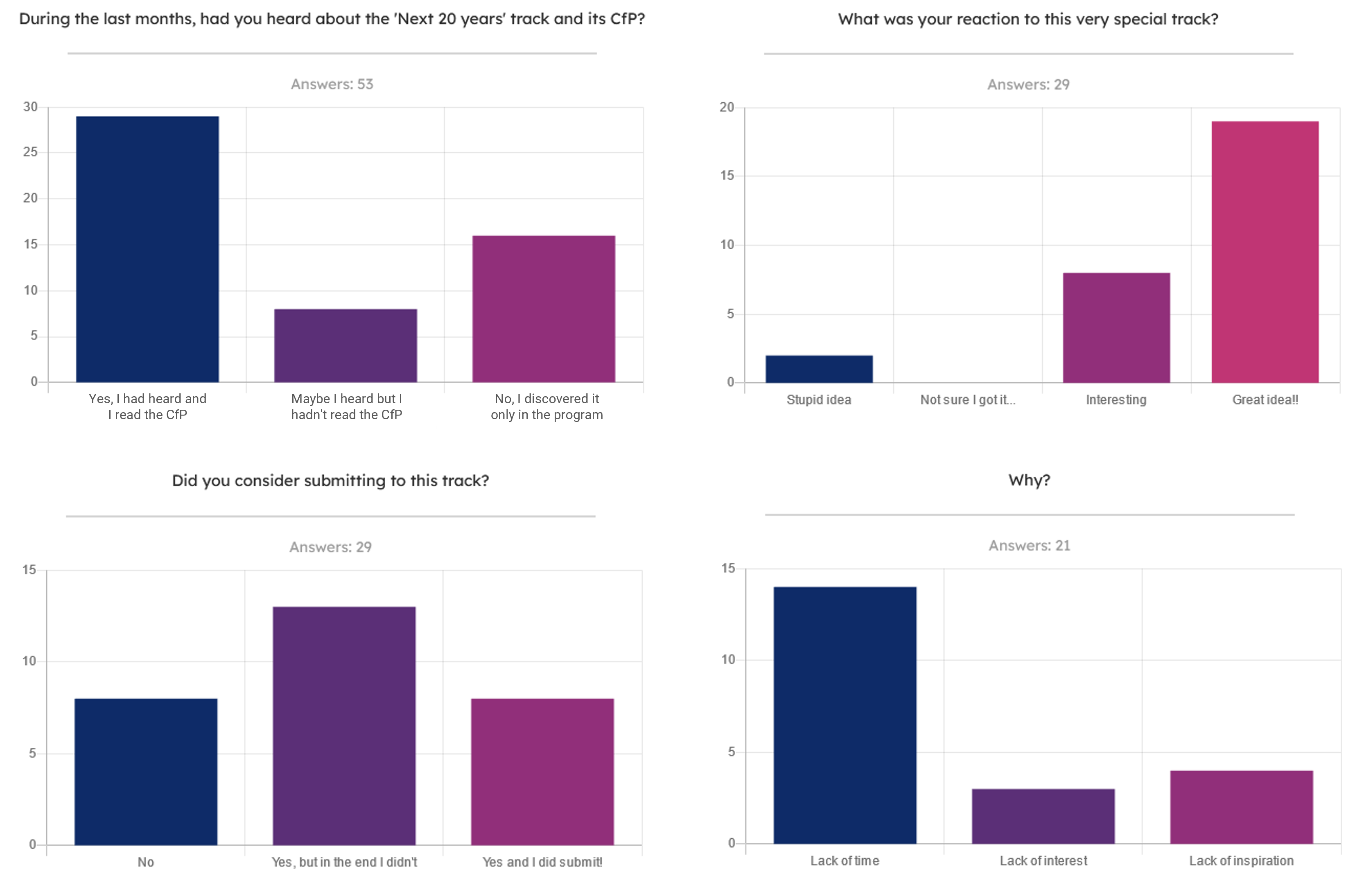}
    \caption{Survey to attendees: questions about the track and its CfP.}
    \label{fig:q-call}
\end{figure}

We first asked some questions about the track, the call for papers and if they considered applying: the results are summarized in Figure~\ref{fig:q-call}. Almost 55\% of respondents had already heard about the track before the conference but 30\% discovered it only in the program (top-left); almost all of those who spotted the CfP thought it was an interesting or great idea (93\%, top-right); a large majority considered to submit or even submitted (74\%, bottom-left), while those who did not submit were mainly discouraged by the lack of time (67\%, bottom-right).

\begin{figure}[ht]
    \centering
    \includegraphics[width=\textwidth]{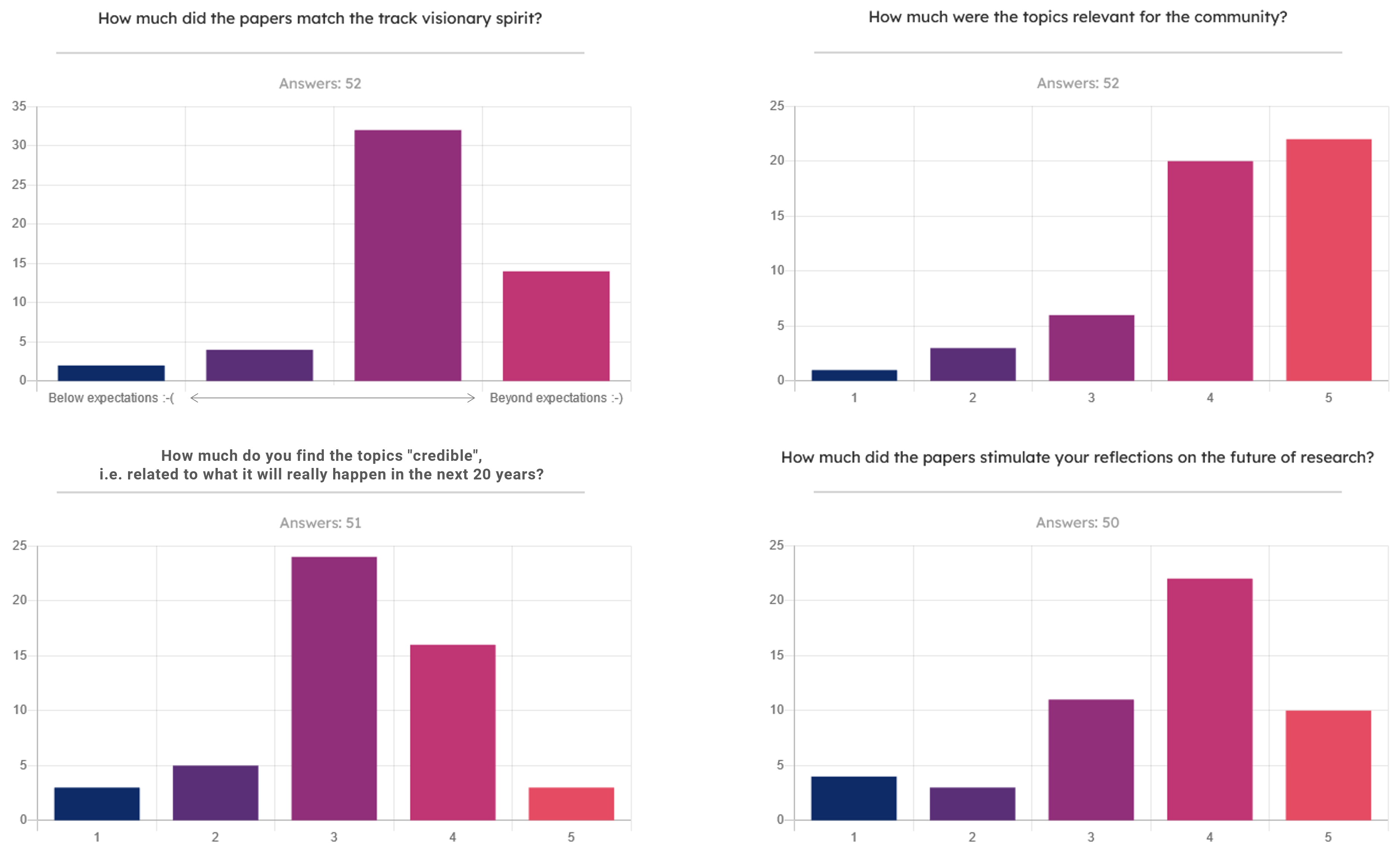}
    \caption{Survey to attendees: questions about the accepted papers and their content.}
    \label{fig:q-papers}
\end{figure}

Then, we investigated the audience opinion about the papers and their contents, as displayed in Figure~\ref{fig:q-papers}. An 88\% majority was satisfied by how the papers matched the visionary spirit of the track, with 27\% who thought that the presented work was beyond their expectation (top-left); the relevance of the paper topics to the community was more questioned, but still 80\% of respondent gave a medium-high or high score (top-right); the perceived credibility of the topics in relation to the next 20 years was much lower, with 47\% giving a medium score (bottom-left), but a large share of respondents was stimulated to reflect on the future of research (64\% with a medium-high or high score, bottom-right). We also asked if there was any research topic that was partially or completely missing in the presented work, and almost 50\% gave some suggestion; the mentioned topics include: graph federation, data capture, ontologies vs. embeddings, reasoning and stream reasoning, dynamic knowledge construction, query processing, constraint validation, access control, future formats, metaverse and virtual worlds, computation at hardware level, misinformation, material science, climate change, health assistants, decision-support systems.

%\textcolor{red}{"Future formats, metaverse and virtual worlds", "Impact of semantic web on climate change, AI alignment", "Imagine a future were data integration is solved with open world computation at hardware level, and the role of semantic community is creating barriers without loosing fairness", "Misinformation, Material science, Health assistants", "Changes in data capture", "I would like to see more about federated learning and edge computing", "climate change", "directly from the Open World assumption, I think there are many topics that were not included :) maybe there could have been more about the federation aspect - how different graphs will 'talk to each other' (maybe the AI agents vision will become reality) - not necessarily as robots per-se, but e.g. as enhanced decision support systems in day-to-day lives. there could also have been some (controversial / thought provoking) discussion on the diminishing role of ontologies in a world of embeddings. will embeddings converge to / replace ontologies? Since they effectively represent the world as it is described in text? and other such topics", "Reasoning and specifically Stream Reasoning", "Dynamic Knowledge Construction over Knowledge Discovery", "Query Processing, Constraint Validation, Access Control"}. % TO BE SUMMARISED!! 

\begin{figure}[ht]
    \centering
    \includegraphics[width=\textwidth]{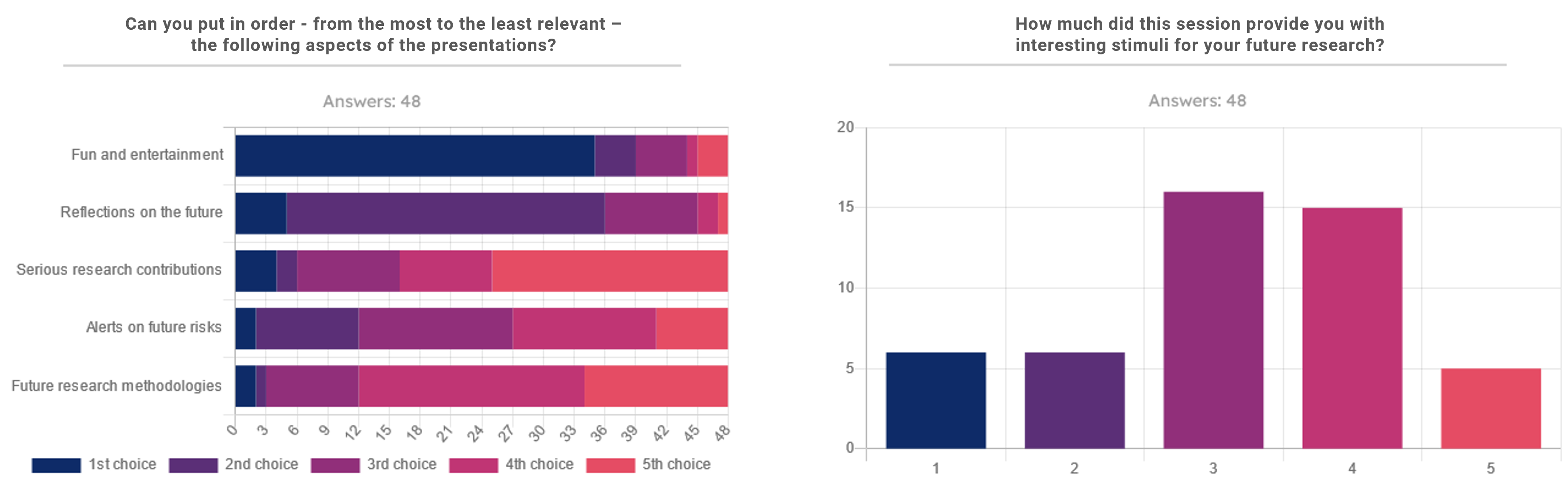}
    \caption{Survey to attendees: questions about the overall session.}
    \label{fig:q-feedback}
\end{figure}

We asked the audience also feedback on the session overall. Regarding the presentations, 50\% liked them a lot and 44\% thought they were great, confirming the general perception that the session was appreciated. Asking the audience to put in order some aspects of the presentations, it is clear that the fun and entertainment angle was the most appreciated one, followed by reflections on the future (cf. left side of Figure~\ref{fig:q-feedback}); the audience answer was quite varied in terms of the stimuli they received for their future research (cf. right side  of Figure~\ref{fig:q-feedback}). 

\begin{figure}[ht]
    \centering
    \includegraphics[width=0.6\textwidth]{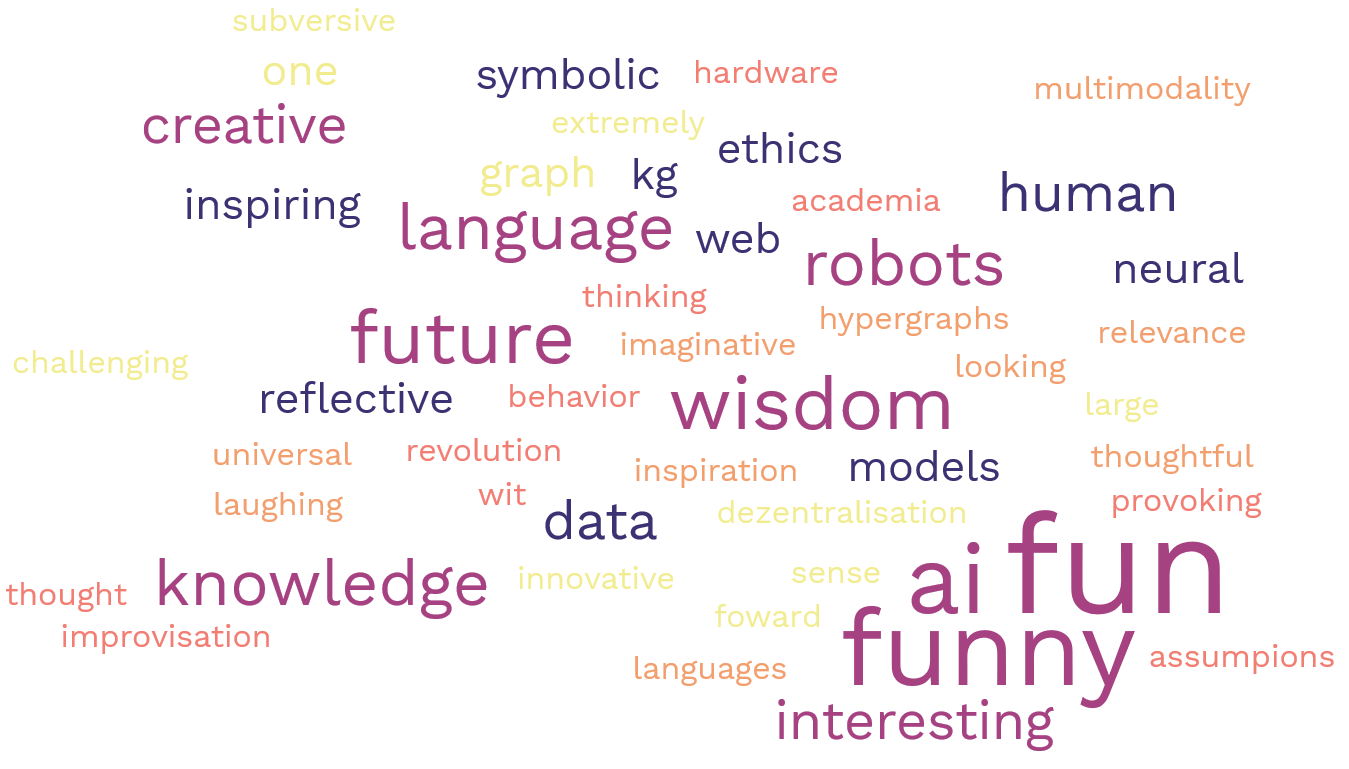}
    \caption{Answers from the free text question}
    \label{fig:word-cloud-free-text}
\end{figure}

Towards the end of the questionnaire, we asked participants a free text question to write three words that came to their mind to sponteneaously characterize the session. The results are depicted in Figure~\ref{fig:word-cloud-free-text}. A lot of the answers emphasized the fun aspect of the session, while there were also some research topics named (AI, language, robots, ...), which shows that the audience did perceive the session from both a research as well as an entertainment angle.

\begin{figure}[ht]
    \centering
    \includegraphics[width=\textwidth]{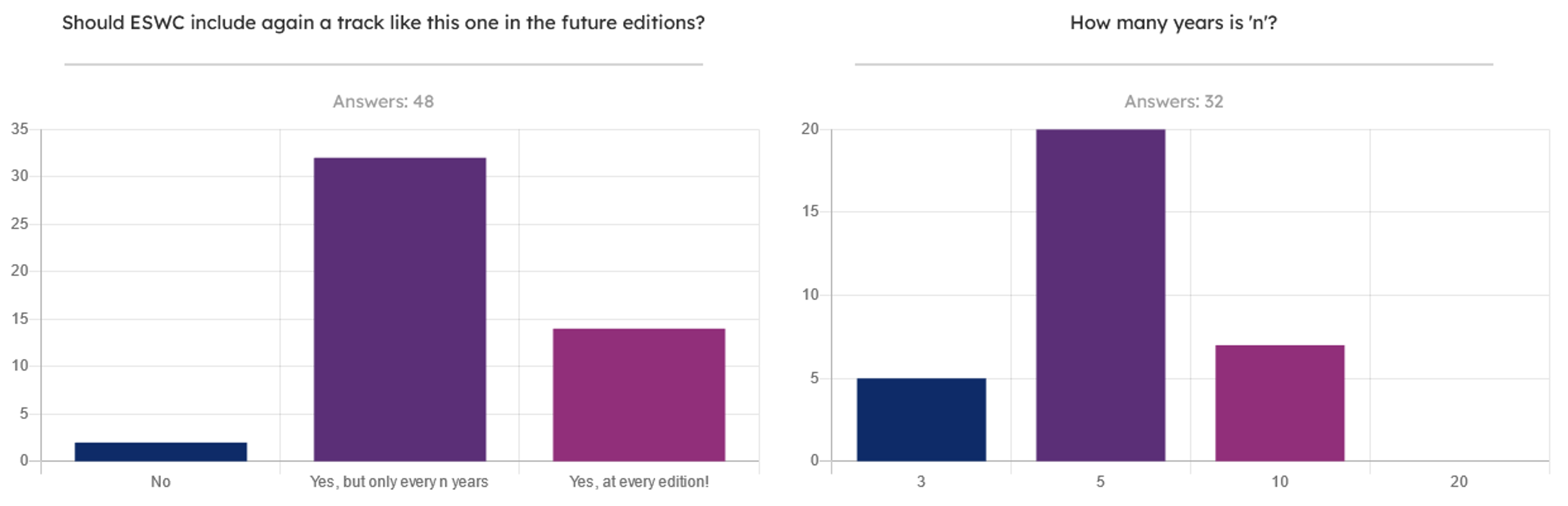}
    \caption{Survey to attendees: questions about potential repetitions of the track.}
    \label{fig:q-future}
\end{figure}

Finally, we checked the interest of the audience with respect to new editions of a similar track. Figure~\ref{fig:q-future} shows on the left that 96\% of respondents would welcome this idea, while on the right we see that a large majority suggests to repeat the track every 5 years.

During the rest of the conference, we also collected other comments and considerations by the conference participants. In general, our impression that attendees either loved the fun and entertaining experience or were partially disappointed by the fact that the balance of the session was too much on the fiction side and too little on the research side: a few people commented that they would have rather preferred a ``serious'' reflection on future Semantic Web visions after the presentations, even if everybody enjoyed the relaxed and creative presentations. One attendee also commented that, with respect to Black Mirror-inspired scientific event he had participated in the past, the presentations and discussions in the session were oriented towards more positive future visions and constructive forecasts, instead towards dystopian or gloomy predictions of risks and problems. 

% other comments by the authors: as an exercise to be repeated as stimulates creativity, as interest for phd student to develop skills, as followup ws to work on the fictions to draw insights and outcomes out of it
In this respect, one of the authors commented that he would have not liked a ``serious'' discussion following the presentations, but he envisioned the possibility of organising follow-up workshops to take those fictional visions and discuss about the concrete future of Semantic Web research. Indeed, our intention as track chairs was to provide a fun and interesting space for discussions about (potentially crazy) ideas for the future, in order to stimulate reflection and to solicit emerging concepts and breakthroughs after the session (rather than during it).

Regarding the potential repetition of this kind of track, we would like also to report the feedback received from the PhD student authors of one of the papers: they underlined the importance of this ``exercise'' to stimulate creativity and to develop important research skills, which is specifically useful for student. As a matter of fact, this has already been considered as part of computer science and AI curricula~\cite{goldsmith2011science,goldsmith2014fiction}, exactly with the goal of exploiting fiction as an introduction to research. This was also mentioned by another author that cited a popular book published in 1997~\cite{stork1997hal} that explored how 2001 Space Odyssey had influenced the research and design of intelligent machines.

%%%%%%%%%%%%%%%%%%%%%%%%%%%%%%%%%%%%%%%%%%%%%%%%%%%%%%%%%%%%%%%%%%%%%%%%%%%%%%%%%%%%%%%%%%%%
\section{Publication of the Proceedings}\label{sec:ceur}
Our initial plan was to publish the proceedings of the paper with CEUR-WS\footnote{\url{https://ceur-ws.org/}}. This, however, lead to a few conflicts with the organization's guidelines for publication. First, CEUR-WS asked for AI co-authors to be removed, as they are currently not allowed for papers published on CEUR-WS.\footnote{\url{https://ceur-ws.org/ACADEMIC-ETHICS.html}}

At a later stage, CEUR-WS pointed to further issues. Essentially, they did not want to publish papers with ``false'' or ``inventend'' contents (where the latter, by design, was the spirit of the track). This held both for the actual papers' contents, as well as for the references sections. Another concern raised was the potential indexing of future papers by engines such as Google scholar, in which case CEUR-WS would have been involved in injecting false information into those engines.

Ultimately, it was mutually agreed that we withdraw the volume from CEUR-WS. The proceedings will now be published on Zenodo~\cite{eswc2043proceedings}. Time will tell whether the future references will be picked up by Google scholar or not.%\footnote{Note to reviewers: we expect further updates and insights here by the time the final version is due.}

%\newpage
%%%%%%%%%%%%%%%%%%%%%%%%%%%%%%%%%%%%%%%%%%%%%%%%%%%%%%%%%%%%%%%%%%%%%%%%%%%%%%%%%%%%%%%%%%%%
\section{In Lieu of Conclusions}\label{sec:concl}
%considerations/speculations about the future of semweb research and research methodologies

% track as a sort of social experiment and not only sneak peek on future research
In this paper, we presented an overview and analysis  of the outcomes of the ``ESWC 2043 - Next 20 Years'' track of the Extended Semantic Web Conference 2023. As track chairs, we had the opportunity to run this sort of social experiment involving the Semantic Web community. Our goals were to solicit an innovative form of reflection on the future of research in this area, to collect visions and "bets" from the community, and of course also to have fun. The results went beyond our expectations, in terms of participation of researchers, content of papers  and session reception by the audience. We believe that this is a clear sign of the vitality of the Semantic Web community.

% interesting reflections on if and how our way to do research may change in the future (and make use of technologies in the research process as well)
Our analysis revealed not only some potential ``sneak peeks'' into the future, but confirmed that the research process itself may change over time, as the authors (with or without consciously reflecting on it) applied a mix of traditional methods and innovative practices. The design fiction characteristics of the call forced them to focus on the (future) storytelling; they employed technologies in various ways to assist their research process, experimenting the recent advancements in generative AI tools as part of the paper writing; they were able to test and check the possibilities and limitations of such technologies, in support to brainstorming, exploration and composition.

% confirmation of main trending topics of the semantic web and things we should remember to have the semweb community still relevant in 20 years <-- MAY CHANGE AFTER HEIKO'S ANALYSIS
Topic-wise, the accepted papers confirmed the main themes and trends of the Semantic Web community, inserting some additional flavours: decentralization and data spaces, personal data management and privacy/control, interplay with other Artificial Intelligence sub-fields, common sense and interaction with natural and synthetic entities, new devices and applications, technology impact on various domains. Most authors indeed made an effort to reflect upon what would make the Semantic Web still relevant in 20 years, whether or not we will still use such name (instead, for example, the ``Wisdom Web'' name proposed in a couple of papers).

% relevance of the design fiction approach: both for phd students to develop skills and for the community just to kickstart a discussion, but maybe worth a followup
The response of the conference audience was also very interesting, in that attendees actively accepted the challenge of the fictional discussion, still conveying relevant content and actual challenges in their questions to the authors' panel. An open challenge, in case of future editions of this kind of track, would be to find a proper balance between the fun/imaginary content and the meaningful debate, to make room both for the design fiction aspect and for a serious conversation on the open issues and future visions.

We still believe that this very special track was totally worth the effort, in that not only it led to very interesting results, but also because the design fiction approach proved to be effective in soliciting researchers to ``think outside the box''. We agree with those authors who stated that this method could be very valuable for PhD students to develop their research skills. We also think that this approach can also be beneficial for a scientific community at large, to kick-start a broader discussion on the future trends and challenges of a discipline. In (lieu of) conclusion, we offer this time traveler's guide to Semantic Web research as an initial seed to promote the growth of a fun yet relevant dialogue on the future research in our community.

\section*{Acknowledgements}

First of all, we want to thank Catia Pesquita -- General Chair of ESWC 2023 -- for assigning us the role of chairs for the ``Next 20 years'' special track and for allowing us to ``stretch'' the scope of the track and experimenting the design fiction approach. We also want to thank all the authors that submitted their papers to this special track and that contributed with their presentations at the conference, as well as all ESWC 2023 attendees who participated to this special session and also responded to our survey.

%\newpage

%%
%% Bibliography
%%

\bibliographystyle{splncs04}
\bibliography{tgdkbib}

\end{document}